\documentclass[11 pt]{article}  

\pdfoutput=1

\usepackage{epsfig} 
\usepackage{times} 
\usepackage{amssymb}
\usepackage{mdframed}
\usepackage{lipsum}
\usepackage{amsfonts}
\usepackage{paralist}
\usepackage{bbding}
\usepackage{pifont}
\usepackage{float}
\usepackage{fancyvrb}
\usepackage[bottom]{footmisc}
\usepackage{wrapfig}
\usepackage{color}
\usepackage[T1]{fontenc}
\usepackage[outline]{contour}
\usepackage{xcolor}
\usepackage{amsmath}
\usepackage{mathtools}

\usepackage{hyperref}
\hypersetup{
	colorlinks=true,
	citecolor={blue},
	linkcolor = {blue},
	bookmarksopen=false,
	bookmarksnumbered=true
}

\usepackage[mathscr]{euscript}
\usepackage{nameref}
\usepackage[vlined]{algorithm2e}
\usepackage{adjustbox}
\usepackage{mathpartir,xparse}
\usepackage{bbm}

\usepackage{listings}
\definecolor{light-blue}{rgb}{0.4,0,0.9}
\newcommand{\starl}[1]{\textcolor{light-blue}{#1}}
\makeatletter 
\def\arcr{\@arraycr}
\makeatother

\usepackage{wasysym}
\usepackage{verbatim}
\usepackage{tikz}
\usetikzlibrary{calc,patterns,decorations.pathmorphing,decorations.markings,shapes, arrows}
\usepackage{fullpage}
\usepackage[font=small,labelfont=bf]{caption}
\usepackage{subcaption}

\newcommand{\tb}[1]{\textbf{#1}}

\newcommand{\lgname}{K\kern-2pt oord\xspace}
\newcommand{\toolname}{K\kern-2pt oordinator\xspace}

\newcommand{\myuin}{\mathit{pid}\xspace}

\makeatletter
\newcommand\notsotiny{\@setfontsize\notsotiny\@vipt\@viipt}
\makeatother

\newcommand{\refsect}[1]{Section~\ref{sec:#1}}
\newcommand{\reffig}[1]{Figure~\ref{fig:#1}}
\newcommand{\reftab}[1]{Table~\ref{tab:#1}}

\definecolor{orange}{rgb}{1,0.5,0}
\definecolor{darkgreen}{rgb}{0.0, 0.5, 0.0}
\newcommand{\marker}[1]{} 

\newcommand{\sayan}[1]{{\textcolor{black}{#1}}}

\newcommand{\rg}[1]{\textcolor{black}{#1}}

	\lstdefinelanguage{xyz}{
		keywordstyle=\bf, 
		identifierstyle=\it, 
		emphstyle=\starl, 
		mathescape=true,
		tabsize=20,
		sensitive=false,
		columns=fullflexible,
		keepspaces=true,
		flexiblecolumns=true,
		basewidth=0.05em,
		moredelim=[il][\rm]{//},
		moredelim=[is][\sf ]{!}{!},
		moredelim=[is][\bf ]{*}{*},
		keywords={
                        and , module,
			agent, sensors, actuators, allread, allwrite, atomic, actuator, assume,
			else,elseif,end,eff,
			atomic,for, foreach,forget,
			input,internal,if,import,in,invariant,
			local,
			or,
			pre,
			return,
			sensor,
			then,def,type,types,thread,to,
			using,module
			variables, vocabulary,
			when,where, with,while},
		emph={update,replaceList,findTask,finishTask,beginTask,doMove, doReach, doReachA, forget, getOdometer, getInput, getMotion, getObs, getPos, getxPos, getPosition, getNeighbors, setMotion, inMotion, sharedsw, sharedmw, tuple, map, array, enumeration},
		literate=
		{(}{{$($}}1
		{)}{{$)$}}1
		{\\in}{{$\in\ $}}1
		{\\preceq}{{$\preceq\ $}}1
		{\\subset}{{$\subset\ $}}1
		{\\subseteq}{{$\subseteq\ $}}1
		{\\supset}{{$\supset\ $}}1
		{\\supseteq}{{$\supseteq\ $}}1
		{\\forall}{{$\forall$}}1
		{\\le}{{$\le\ $}}1
		{\\ge}{{$\ge\ $}}1
		{\\gets}{{$\gets\ $}}1
		{\\cup}{{$\cup\ $}}1
		{\\cap}{{$\cap\ $}}1
		{\\langle}{{$\langle$}}1
		{\\rangle}{{$\rangle$}}1
		{\\exists}{{$\exists\ $}}1
		{\\bot}{{$\bot$}}1
		{\\rip}{{$\rip$}}1
		{\\emptyset}{{$\emptyset$}}1
		{\\notin}{{$\notin\ $}}1
		{\\not\\exists}{{$\not\exists\ $}}1
		{\\ne}{{$\ne\ $}}1
		{\\to}{{$\to\ $}}1
		{\\implies}{{$\implies\ $}}1
		{<*>}{{$\langle*\rangle$}}1
		{=}{{$=\ $}}1
		{~}{{$\neg\ $}}1
		{|}{{$\mid$}}1
		{'}{{$^\prime$}}1
		{\\A}{{$\forall\ $}}1
		{\\E}{{$\exists\ $}}1
		{\\/}{{$\vee\,$}}1
		{\\vee}{{$\vee\,$}}1
		{/\\}{{$\wedge\,$}}1
		{\\wedge}{{$\wedge\,$}}1
		{=>}{{$\Rightarrow\ $}}1
		{->}{{$\rightarrow\ $}}1
		{<=}{{$\ \leq\ $}}1
		{<-}{{$\leftarrow\ $}}1
		{==}{{$=\mathrel{\mkern-3mu}=\ $}}1
		{_}{{\_}}1
		{~=}{{$\neq\ $}}1
		{\\U}{{$\cup\ $}}1
		{\\I}{{$\cap\ $}}1
		{|-}{{$\vdash\ $}}1
		{-|}{{$\dashv\ $}}1
		{<<}{{$\ll\ $}}2
		{>>}{{$\gg\ $}}2
		{||}{{$\|$}}1
		{[[}{{$\langle$}}1
		{]]}{{$\rangle$}}1
		{[}{{$[$}}1
		{]}{{$\,]$}}1
		{]]]}{{$]\rangle$}}1
		{<=>}{{$\Leftrightarrow\ $}}2
		{<->}{{$\leftrightarrow\ $}}2
		{(+)}{{$\oplus\ $}}1
		{(-)}{{$\ominus\ $}}1
		{_i}{{$_{i}$}}1
		{_j}{{$_{j}$}}1
		{_{i,j}}{{$_{i,j}$}}3
		{_{j,i}}{{$_{j,i}$}}3
		{_0}{{$_0$}}1
		{_1}{{$_1$}}1
		{_2}{{$_2$}}1
		{_n}{{$_n$}}1
		{_p}{{$_p$}}1
		{_k}{{$_n$}}1
		{@}{{}}0
		{\\delta}{{$\delta$}}1
		{\\R}{{$\R$}}1
		{\\Rplus}{{$\Rplus$}}1
		{\\N}{{$\N$}}1
		{\\times}{{$\times\ $}}1
		{\\tau}{{$\tau$}}1
		{\\alpha}{{$\alpha$}}1
		{\\beta}{{$\beta$}}1
		{\\gamma}{{$\gamma$}}1
		{\\ell}{{$\ell\ $}}1
		{--}{{$-\ $}}1
		{\\TT}{{\hspace{1.5em}}}3
	}
	
	\lstdefinelanguage{xyzNums}[]{xyz}
	{
		numbers=left,
		numbersep=15pt,
		numberstyle=\tiny,
		stepnumber=1,
		numbersep=4pt,
		xleftmargin=2em,
		frame=single,
	}
	
	\lstdefinelanguage{xyzNumsRight}[]{xyz}
	{
		numbers=right,
		numbersep=15pt,
		numberstyle=\tiny,
		stepnumber=1,
		numbersep=4pt,
		xleftmargin=2em,
		frame=single,
		framexrightmargin=-1.5em,
		framexleftmargin=1.5em
	}

\newcommand{\num}[1]{\relax\ifmmode \mathbb #1\else $\mathbb #1$\fi}
\newcommand{\nnnum}[1]{\relax\ifmmode 
	{\mathbb #1}_{\geq 0} \else ${\mathbb #1}_{\geq 0}$
	\fi}
\newcommand{\npnum}[1]{\relax\ifmmode 
	{\mathbb #1}_{\leq 0} \else ${\mathbb #1}_{\leq 0}$
	\fi}
\newcommand{\pnum}[1]{\relax\ifmmode 
	{\mathbb #1}_{> 0} \else ${\mathbb #1}_{> 0}$
	\fi}
\newcommand{\nnum}[1]{\relax\ifmmode 
	{\mathbb #1}_{< 0} \else ${\mathbb #1}_{< 0}$
	\fi}
\newcommand{\plnum}[1]{\relax\ifmmode 
	{\mathbb #1}_{+} \else ${\mathbb #1}_{+}$
	\fi}
\newcommand{\nenum}[1]{\relax\ifmmode 
	{\mathbb #1}_{-} \else ${\mathbb #1}_{-}$
	\fi}

\newcommand{\reals}{{\num R}}                    

\definecolor{WowColor}{rgb}{.75,0,.75}
\definecolor{SubtleColor}{rgb}{0,0,.50}
\definecolor{TODOColor}{rgb}{0,.50,0}

\newcounter{margincounter}

\newcommand{\fTBD}[1]{}

\newcommand{\Gazebo}{Gazebo\xspace}
\newcommand{\Koord}{\ensuremath{\lgname}\xspace}

\newcommand{\Lineform}{\textit{Lineform}\xspace}
\newcommand{\Shapeform}{\textit{Shapeform}\xspace}
\newcommand{\TaskApp}{\textit{Task}\xspace}
\newcommand{\CyPhyHouse}{\textit{CyPhyHouse}\xspace}
\AtBeginDocument{
  \setlength\abovedisplayskip{0pt}
  \setlength\belowdisplayskip{0pt}}

\title{\bf CyPhyHouse: A Programming, Simulation, and Deployment Toolchain for Heterogeneous Distributed Coordination}

\author{Ritwika Ghosh$^{1}$, Joao P. Jansch-Porto$^{2}$, Chiao Hsieh$^{1}$, Amelia Gosse$^{2}$, \\ Minghao Jiang$^{3}$, Hebron Taylor$^{2}$, Peter Du$^{3}$, Sayan Mitra$^{3}$, Geir Dullerud$^{2}$
\thanks{Coordinated Science Laboratory, University of Illinois at Urbana-Champaign, Urbana, IL 61801}
\thanks{\textsuperscript{1} Department of Computer Science}
\thanks{\textsuperscript{2} Department of Mechanical Science and Engineering}
\thanks{\textsuperscript{3} Department of Electrical and Computer Engineering}
\thanks{\texttt{\{rghosh9,janschp2,chsieh16,gosse2,mjiang24,hdt2,peterdu2,mitras,dullerud\} @illinois.edu}}
}

\renewcommand\footnotemark{}

\begin{document}
\maketitle


\begin{abstract}
Programming languages, libraries, and development tools have transformed the application development processes for mobile computing and machine learning. 
This paper introduces the \CyPhyHouse---a toolchain that aims to provide similar programming, debugging, and deployment benefits for distributed mobile robotic applications. Users can develop hardware-agnostic, distributed applications using the high-level, event driven \Koord programming language, without requiring expertise in controller design  or distributed network protocols. The  modular, platform-independent \emph{middleware} of \CyPhyHouse implements these functionalities using standard algorithms for path planning (RRT), control (MPC), mutual exclusion, etc. A  high-fidelity, scalable, multi-threaded simulator for \Koord applications is developed to  simulate the same application code for dozens of heterogeneous agents. The same compiled code can also be  deployed on heterogeneous mobile platforms. The effectiveness of \CyPhyHouse in improving the design cycles is explicitly illustrated in a robotic testbed through development, simulation, and deployment of a distributed task allocation application on in-house ground and aerial vehicles.
\end{abstract}

\section{Introduction}
\label{sec:intro}
Programming languages like C\#, Swift, Python, and development tools like LLVM~\cite{llvm} have helped make millions of people, with diverse backgrounds, into mobile application developers. 
Open source software libraries like Caffe~\cite{jia2014caffe}, PyTorch~\cite{paszke2017automatic} and Tensorflow~\cite{tensorflow2015-whitepaper} have propelled the surge in machine learning research and development. To a lesser degree, similar efforts are afoot in democratizing robotics. Most prominently,  ROS~\cite{ros} provides hardware abstractions, device drivers, messaging protocols, many common library functions and has become widely used. The PyRobot~\cite{pyrobot2019} and PythonRobotics~\cite{PythonRobotics} libraries provide hardware-independent implementations of common functions for physical manipulation and navigation of individual robots.

%
 
Nevertheless, it requires significant effort (weeks, not hours) to develop, simulate, and debug a new application for a single mobile robot---not including the effort to build the robot hardware. The required effort grows quickly for distributed and heterogeneous systems, as none of the existing robotics libraries provide either \begin{inparaenum}[(a)]\item support for distributed  coordination, or \item easy portability of code across different platforms.\end{inparaenum} 

 With the aim of simplifying application development for distributed and heterogeneous systems, in this paper we introduce  \CyPhyHouse\footnote{\href{https://cyphyhouse.github.io/index.html}{https://cyphyhouse.github.io}}---an open source software toolchain for programming, simulating, and  deploying mobile robotic applications.

In this work, we target distributed coordination tasks such as collaborative  mapping~\cite{cunningham2010ddf}, surveillance, delivery, formation-flight, etc. with aerial drones and ground vehicles. We believe that for these applications, low-level motion control for the individual robots is \sayan{standard but tedious}, and coordination across distributed (and possibly heterogeneous) robots is particularly difficult and error-prone. This motivates the two key abstractions provided by \CyPhyHouse: (a) {\em portability\/} of high-level coordination code across different platforms; and (b) {\em shared variable\/} communication across robots.

%
%
%



The first of the several software components of \CyPhyHouse is a high-level programming language called \Koord that enables users to write distributed coordination applications without being encumbered by socket programming, ROS message handling, and thread management.
Our \Koord compiler generates code that can be and has been directly deployed on aerial and ground vehicle platforms as well as simulated with the {\em \CyPhyHouse simulator.}
%
%
\Koord language abstractions for path planning, localization, and shared memory make application programs succinct, portable, and readable (see~\refsect{overview}).
The modular structure
 of the \CyPhyHouse\ {\em middleware\/} we have built will make it easy for a roboticist to add support for new hardware platforms. 
In summary, the three main contributions of this paper are as follows.
 \fTBD{\sayan{What other high-level claims are we making?}}

%

%

 \begin{figure*}[h!] 
    \centering
    \includegraphics[width=\textwidth]{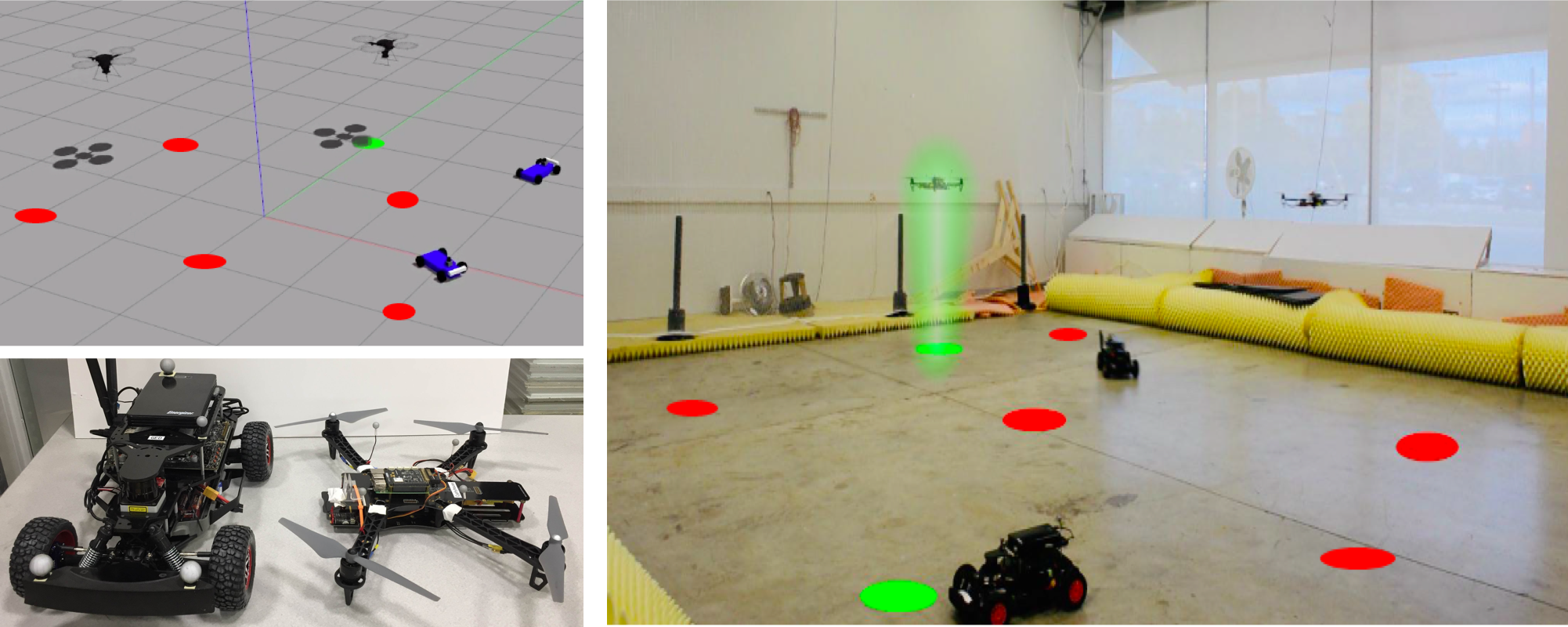}
    \vspace{0.1cm}
 \caption{\emph{Right:} Annotated snapshot of a distributed task allocation application deployed on four cars and drones using \CyPhyHouse in our test arena. The red tasks are incomplete, and the green are completed. {\em Left bottom:\/}  different robotic platforms: the F1/10 Car and the quadcopter.  \emph{Left top:\/} Visualization of the same application running in \CyPhyHouse simulator which interfaces with $\Gazebo$.}
  \label{fig:realvis}
\end{figure*}

\paragraph*{1. An end-to-end distributed application for robotic vehicles and drones developed and deployed using \CyPhyHouse toolchain}
This \TaskApp application requires the participating robots to visit a common list of points, mutually exclusively, and while avoiding collisions.
Our solution program written in \Koord is less than 50 lines long (see \reffig{taskoutline}). Our compiler generates executables for both the drone and the vehicle platforms, linking the platform independent parts of the application with the platform-specific path planners and controller. 
We ran more than 100 experiments with a set of tasks running on different combinations of ground and aerial vehicles, all with few edits in the {\em configuration file\/} (see \reffig{config}).
 

\paragraph*{2. A high-fidelity, scalable, and flexible simulator for distributed heterogeneous systems}
The simulator executes instances of the  application code generated by the \Koord compiler---one for each robot in the scenario. Within the simulator, individual robots communicate with each other over a wired or a wireless network and with their own simulated sensors and actuators through ROS topics. For example, a simulation with 16 drones can spawn over 1.4K ROS topics and 1.6K threads, yet, our simulator is engineered to execute and visualize such scenarios in \Gazebo running on  standard workstations and  laptops. In Section~\ref{sec:simulator}, we present detailed performance analysis of the simulator. \fTBD{mention realtime factor?}

\paragraph*{3. A programming language and middleware for heterogeneous platforms that support application development, simulation, deployment, as well as verification\footnote{Formal semantics of the language and the automatic verification tools are not part of this paper. Some of the details of the formal aspects of \Koord were presented in an earlier workshop paper~\cite{koordwshop}.}} As explained earlier, \Koord provides abstractions for distributed applications running on possibly heterogeneous platforms. For example, \Koord supports easy coordination across robots: a single line of code like
\[
x[\myuin] = (x[\myuin-1] + x[\myuin+1])/2
\]
assigns to a shared variable $x[\myuin]$ of a robot with the unique integer identifier $\myuin$, the average of the values of $x[\myuin-1]$ and $x[\myuin+1]$ which are the values held respectively by robots $\myuin-1$ and $\myuin+1$.
This makes \Koord implementations of consensus-like protocols read almost like their textbook counterparts~\cite{Magnusbook2010}.
These statements are implemented using message-passing in the \CyPhyHouse middleware.
\Koord comes with well-defined semantics which makes it possible to reason about the correctness of the distributed applications using formal techniques.
\fTBD{\sayan{Finish with something stronger.. ``to our knowledge, this is the first ...''. }}



\section{Related work}
\label{sec:related} 
Several frameworks and tools address the challenges in development of distributed robotic applications. \reftab{summary} compares these works along the following  dimensions: 
\begin{inparaenum}[(a)] 
\item whether the framework has been tested with {\bf h}ard{\bf w}are {\bf depl}oyments, 
\item availability of support for networked and {\bf dist}ributed robotic {\bf sys}tems,
\item  support for {\bf heterogeneous} platforms, 
\item availability of specialized {\bf prog}ramming {\bf lang}uage, 
\item availability of a {\bf sim}ulator and {\bf comp}iler, and
\item support for formal {\bf v}erification and {\bf v}alidation.
\end{inparaenum}

Drona provides a language with a decentralized motion planner and builds a mail delivery system (similar to our \TaskApp). The key differences are that Drona uses an asynchronous model of computation (\CyPhyHouse uses a synchronous model) and currently only \CyPhyHouse has demonstrated deployment on heterogeneous platforms.

  Buzz, the programming language used by ROSBuzz~\cite{ROSBuzz} doesn't provide abstractions like CyPhyHouse does with  \Koord, for path planning, de-conflicting, and shared variables. Additionally, ROSBuzz specifically requires the Buzz Virtual Machine to be deployed on each robot platform whereas with \CyPhyHouse, deploying \Koord only requires standard ROS and Python packages.

%
%
%
It should also be mentioned, that  
``Correct-by-construction'' synthesis from high-level temporal logic specifications have been widely discussed in the context of mobile robotics (see, for example~\cite{kress2009temporal,kloetzer2008fully,wongpiromsarn2010receding,wongpiromsarn2011tulip,ulusoy2013optimality}).
\CyPhyHouse differs in the basic assumption that roboticist's (programmer's) creativity and efforts will be necessary well beyond writing high-level specs in solving distributed robotics problems; consequently only the tedious parts of coordination and control are automated and abstracted in the \Koord language and compiler.
\setlength{\tabcolsep}{1.3pt}

\begin{table}[!ht]		
\footnotesize
 \centering		
 \caption{}
 \label{tab:summary}		
   \begin{tabular}{|l| c c c c c c c|}
   	\hline
   	                                     & \tb{HW}    & \tb{Dist.} & \tb{Hetero-} & \tb{Sim}   & \tb{Prog.}         & \tb{Comp.} & \tb{V\&V}  \\
   	\tb{Name}                            & \tb{Depl.} & \tb{Sys.}  & \tb{geneous} &            & \tb{Lang.}         &            &            \\ \hline
   	ROS~\cite{ros}                       & \checkmark &            & \checkmark   & \checkmark & C++/Python/...     &            &            \\
   	ROSBuzz~\cite{ROSBuzz}               & \checkmark & \checkmark & \checkmark   & \checkmark & Buzz               & \checkmark &            \\
   	PythonRobotics &            &            & \checkmark   & \checkmark & Python             &            &            \\
   	PyRobot~\cite{pyrobot2019}           & \checkmark &            & \checkmark  &\checkmark & Python                    &            &            \\
   	MRPT~\cite{MRPT}                     & \checkmark &            & \checkmark   &            & C++                &            &            \\
   	Robotarium~\cite{robotarium}         & \checkmark &            & \checkmark   & \checkmark & Matlab             &            &            \\
   	Drona~\cite{desai2017drona}          & \checkmark & \checkmark &              & \checkmark & P~\cite{Planguage} & \checkmark & \checkmark \\
   	Live~\cite{campusanofabry:lrp2016}   & \checkmark &            & \checkmark   &            & LPR                & \checkmark &            \\
   	CyPhyHouse                           & \checkmark & \checkmark & \checkmark   & \checkmark & \Koord             & \checkmark & \checkmark \\ \hline
   \end{tabular}
 \end{table}

Other open and portable languages that raise the level of abstraction for robotic systems include ~\cite{Bohrer:2018:VVC:3192366.3192406,reactlang,williams2003model}. (For an earlier survey  see~\cite{Nordmann2014}. Most of these older languages are proprietary and  platform-specific.)
%
%
VeriPhy~\cite{Bohrer:2018:VVC:3192366.3192406} also has some commonality with \CyPhyHouse; however, instead of a programming language, the starting point is differential dynamic logic~\cite{Bohrer:2017:FVD:3018610.3018616}.

\section{A distributed task allocation application}
\label{sec:overview}

In this section, we introduce the {\em distributed task allocation problem (\TaskApp) \/} that we will use throughout the paper to illustrate the capabilities of \CyPhyHouse.

Given a robot $G$, and a point $x$ in $R^3$, we say that $G$ has visited $x$ if the position of $G$ stays  within an $\epsilon_v$-ball $x$ for $\delta_v$ amount of time, for some fixed $\epsilon_v>0$ and $\delta_v>0$. The distributed task allocation problem requires a set of robots to visit a sequence of points mutually exclusively:

\begin{quote}
{\small \em \TaskApp: Given a set of (possibly heterogeneous) robots, a safety distance $d_s>0$, and a sequence of points (tasks) $\mathit{list} = x_1, x_2, \ldots \in \reals^3$, it is required that: (a) every unvisited $x_i$ in the sequence is {\em visited\/} exactly by one robot; and (b) no two robots ever get closer than $d_s$.\/}
\end{quote}

We view visiting points as an abstraction for location-based objectives like package delivery, mapping, surveillance, or fire-fighting.

\begin{figure*}[h!]
\centering
    \begin{minipage}[c]{0.2\textwidth}
        \includegraphics[width=\textwidth]{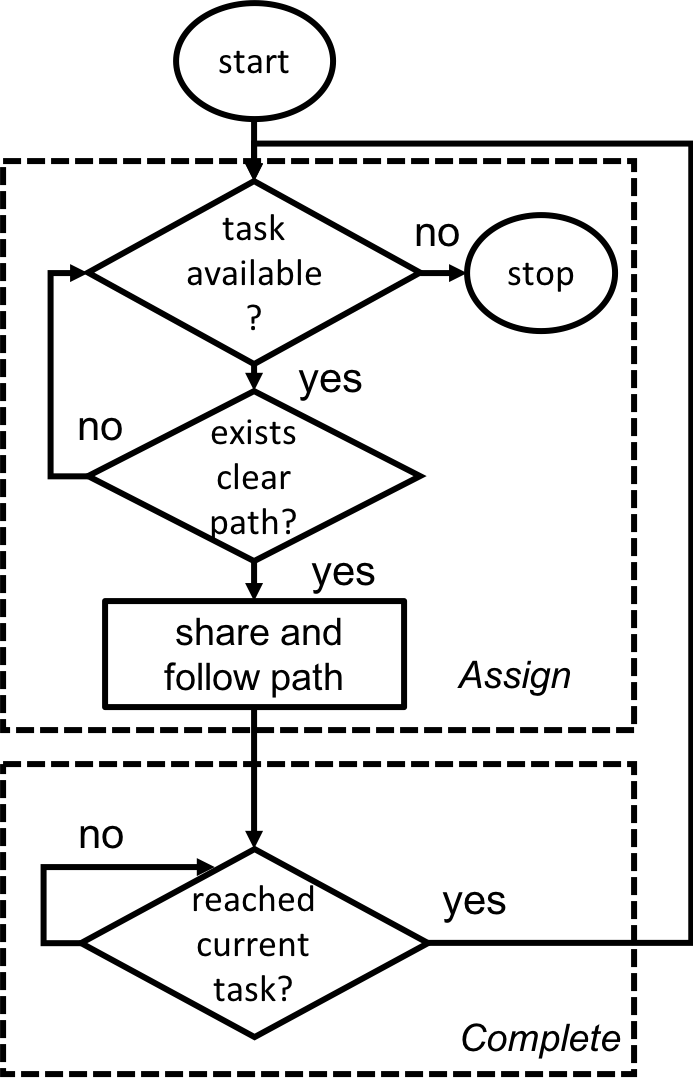}
    \end{minipage}
    \hspace{5pt}
\fbox{
        \notsotiny
        \lstset{language=xyzNums,aboveskip=0em,frame=none}
    \begin{minipage}[c]{0.17\textwidth}
\begin{lstlisting}[firstline=1,lastline=20]
using Motion:
  actuators:
    pos[] path
  sensors:
    pos psn
    boolean reached
local:
  boolean isOnTask
  Task myTask
  pos[] testRoute
  int i
    
allread:
  pos[] route
allwrite:
  Task[] list

Assign:
  pre: isOnTask == false
  eff: if allAssigned(list): stop
\end{lstlisting}
    \end{minipage}
    \hspace{5pt}
    \begin{minipage}[c]{0.26\textwidth}
        {
\begin{lstlisting}[firstline=21,firstnumber=21]
    else: atomic:
      for i = 0, length(list):
        if task.assigned$==$false:
          myTask$=$list[i]
            testRoute$=$findPath(Motion.psn,
                                 myTask.loc)
            if pathIsClear(route, testRoute, pid):
              isOnTask$=$true
              assign(myTask, pid)
              list[i]$=$myTask
              route[pid]$=$testRoute
              Motion.path$=$testRoute
            else:
              isOnTask$=$false
              route[pid]$=$[Motion.psn]
              
Complete:
    pre: isOnTask and Motion.reached
    eff: isOnTask$=$false
         route[pid]$=$[Motion.psn]
         
\end{lstlisting}}
    \end{minipage}}
    \hspace{5pt}
    \begin{minipage}[r]{0.24\textwidth}
    \begin{lstlisting}[basicstyle=\notsotiny\ttfamily, frame=single]
robot:
    pid: 0
    on_device: hotdec_car
    motion_automaton: MotTestCar
    ...
device:
    bot_name: hotdec_car
    bot_type: CAR
    planner: RRT_CAR
    positioning_topic:
        topic: vrpn_client_node/
        type: PoseStamped
    reached_topic: 
        topic: reached
        type: String
    waypoint_topic: 
        topic: waypoint
        type: String
    ...
num_robots: 3
    \end{lstlisting}
    \end{minipage}

    \vspace{0.1cm}
    \caption{
    \emph{Left}\label{fig:taskoutline}
        shows the flowchart for a simple solution to \TaskApp application.
    \emph{Middle}\label{fig:taskapp}
        shows the \TaskApp program implemented in \Koord language for robots with identifier $\myuin$ to solve distributed task allocation problem.
    \emph{Right}\label{fig:config}
        shows snippet of a sample configuration.
        It includes platform-agnostic settings for the robot, e.g., robot id~(\emph{pid}), device to run on~(\emph{on\_device}), and the number of robots (\emph{num\_robots}),
        as well as platform specific settings, e.g., path planners~(\emph{planner}) and position systems~(\emph{positioning\_topic}).
    }
\end{figure*}

The flowchart in \reffig{taskoutline} shows a simple idea for solving this problem for a single robot: Robot $A$ looks for an unassigned task $\tau$ from $\mathit{list}$; if there is a clear path to $\tau$  then $A$ assigns itself the task $\tau$. Then $A$ visits $\tau$ following the path; once done it repeats.
Of course, converting this to a  working solution for a distributed system is challenging as it involves combining distributed mutual exclusion for assigning a task $\tau$ exclusively to a robot $A$ from the $\mathit{list}$ (step 1)~\cite{Lynchbook,ghoshbook},
dynamic conflict-free path planning (step 2), and low-level motion control~(step 3).

Our \Koord language implementation of this flowchart is shown in Figure~\ref{fig:taskapp}. 
It has two {\em events\/}: \emph{Assign} and \emph{Complete}.
The semantics of \Koord is such that execution of the application programs in the distributed system advances in {\em rounds\/} of duration $\delta$\footnote{$\delta$ is a parameter set by the user, with a default value of 0.1 second.}, and in each round, each robot executes at most one event.
A robot can only execute the statements  in the event's effect ({\bf eff}) if its precondition ({\bf pre}) is satisfied. If no event is enabled, the robot does nothing. 
In between the rounds, the robots may continue to move as driven by their local controllers. 
The \Koord middleware (Section~\ref{sec:middleware}) ensures that the robot program executions adhere to this schedule even if local clocks are not precisely synchronized.



In our example, the $\mathit{Assign}$ event uses a single \emph{atomic}  update to assign a task to robot $i$ from the  {\em shared list\/} of tasks 
called $\mathit{list}$ in a mutually  exclusive fashion.\footnote{We provide several library functions associated with abstract data types (assign, allAssigned) and path planners (findPath, pathIsClear). Users can also write functions permitted by \Koord syntax.}
The \emph{route} variable shares paths and positions among all robots, and is used by each robot in computing a collision-free path to an unassigned task.\footnote{Platform specific path-planners can ensure that ground vehicles do not find paths to points above the plane, and aerial vehicles do not find paths to points on the ground.}
To access variables, e.g., \emph{route}, shared by a certain robot, each robot program also has access to its unique integer identifier $\myuin$ and knows the $\myuin$s of all participating robots.

The low-level control of the robot platform is abstracted from the  programmers in \Koord, with certain assurances about the controllers from the platform developers (discussed in \refsect{middleware} and \refsect{hardware}).
The \TaskApp program uses a controller called $\mathit{Motion}$ to drive the robots through a route, as directed by the position value set at its actuator port $\mathit{Motion.route}$.
The sensor ports used by the \TaskApp program are:
\begin{inparaenum}[(a)]
    \item $\mathit{Motion.psn}$: the robot position in a fixed coordinate system.
    \item $\mathit{Motion.reached}$: a flag indicating whether the robot has reached its waypoint.%
\end{inparaenum}

Here the motion module implements vehicle models for the robots. In the next section, we discuss the \CyPhyHouse middleware, which implements a modular design of this runtime system to allow a high degree of flexibility concerning these modules in deployment and simulation.

\section{CyPhyHouse Architecture}
\label{sec:middleware}

A system running a \Koord application has three parts: an application program, a controller, and a plant.
At runtime, the \Koord program executes within the runtime system of a single agent, or a collection of programs execute on different agents that communicate using shared variables.
The plant consists of the hardware platforms of the participating agents.
The controller receives inputs from the program (through actuator ports), sends outputs back to the program (through sensor ports), and interfaces with the plant.
We developed a software-hardware interface (\emph{middleware}) in Python~3.5 to support the three-plane architecture comprising the \Koord runtime system.
\begin{figure}[H]
    \centering
    \includegraphics[width=0.6\textwidth]{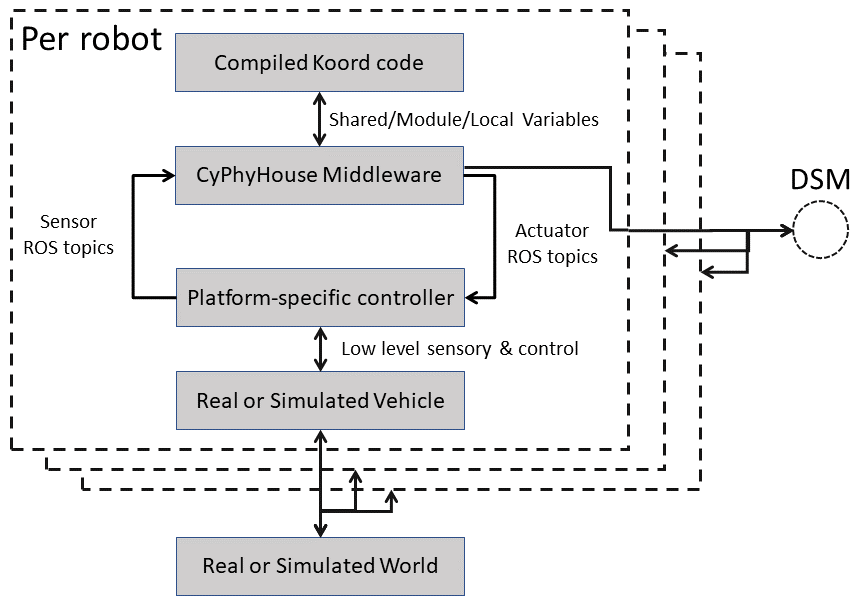}
    \caption{Each compiled \Koord program interacts with \CyPhyHouse middleware simply via variables.
        The middleware implements distributed shared memory(DSM) across agents and the language abstractions over platform-specific controllers through actuator ROS topics,
        and obtain (real or simulated) information such as device positions through sensor ROS topics.\vspace{-2mm}}
    \label{fig:arch1}
\end{figure}

\subsection{Compilation}

The \rg{\Koord compiler included with \CyPhyHouse} generates Python code for the application using all the supported libraries,
such as the implementation of distributed shared variables using message passing over WiFi, motion automata of the robots, high-level collision and obstacle avoidance strategies, etc.
The application then runs with the Python \emph{middleware} for \CyPhyHouse.
The \Koord compiler is written using Antlr (Antlr 4.7.2) in Java~\cite{Parr:2013:DAR:2501720}.%
\footnote{Details of the grammar, AST, and IR design of the \Koord compiler are beyond the scope of this paper;
however, description of the language and its full grammar are provided in~\cite{koordreport} for the interested reader.}
We use ROS to handle the low-level interfaces with hardware.
To communicate between the high-level programs and low-level controllers,
we use Rospy, a Python client library for ROS which enables the (Python) middleware to interface with ROS Topics and Services used for deployment or simulation.

\subsection{Shared memory and Communication}

At a high level, updates to a shared variable by one agent are propagated by the \CyPhyHouse middleware, and become visible to other agents in the next round. The correctness of a program relies on agents having consistent values of shared variables. When an agent updates a shared variable, the middleware uses message passing to inform the other agents of the change. These changes should occur before the next round of computations. 

\CyPhyHouse supports UDP based messaging over Wi-Fi for communication between robots to implement the shared memory. Any shared memory update translates to a update message which the agent broadcasts over WiFi.\footnote{The interested reader is referred to \cite{koordreport} for more details on the shared memory model, and its formal semantics.} The agents running a single distributed \Koord application are assumed to be running on a single network node, with little to no packet loss. However, the communication component of the middleware can be easily extended to support multi-hop networks as well.
\begin{figure*}[h!]
\centering
\begin{subfigure}[b]{0.49\textwidth}
	\centering
	\includegraphics[width=\textwidth]{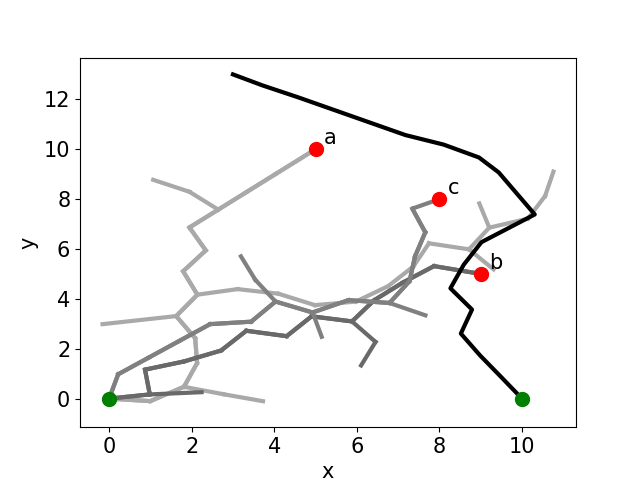}
	\caption{No conflict.}
	\label{subfig:nonconflict}
\end{subfigure}
\begin{subfigure}[b]{0.49\textwidth}
	\centering
	\includegraphics[width=\textwidth]{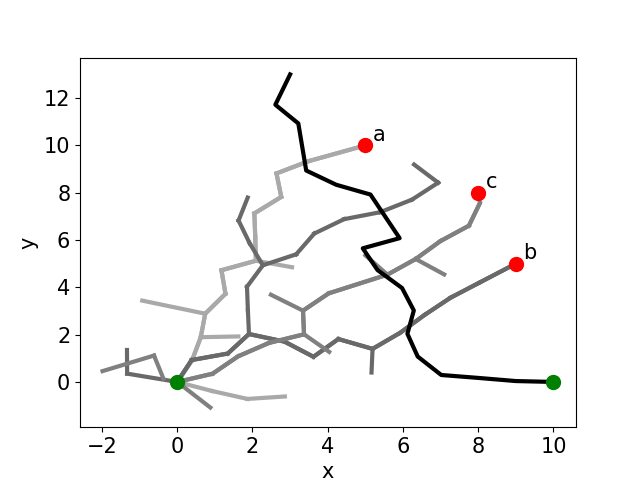}
	\caption{Conflict.}
	\label{subfig:conflict}
\end{subfigure}
\begin{subfigure}[b]{0.49\textwidth}
	\centering
	\includegraphics[width=\textwidth]{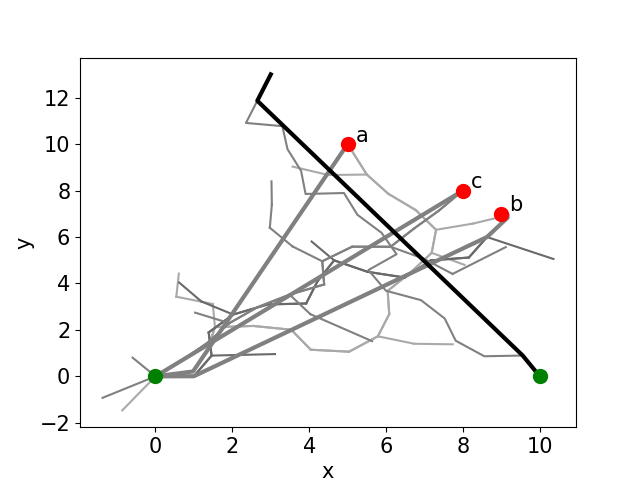}
	\caption{RRT with smoothing.}
	\label{subfig:rrt}
\end{subfigure}
 \caption{Different planner can work with the same code. \textbf{(a)} shows the $x\-y$ plots of concurrently available paths  during a round of the \TaskApp application using an RRT planner for two quadcopters. \textbf{(b)} shows the same configuration, where paths computed are not viable to be traversed concurrently. The green markers are current quadcopter positions, The black path is a fixed path, and the red points are unsassigned task locations. \textbf{(c)} shows the same scenarios under which paths cannot be traversed concurrently, except that a different RRT-based planner (with path smoothing) is used.}
    \label{fig:pathplanners}
\end{figure*}

\subsection{Dynamics}
If an application requires the agents to move, each agent uses an abstract class, \emph{Motion automaton}, which must be implemented for each hardware model (either in deployment or simulation).
This automaton subscribes to the required ROS Topics for positioning information of an agent, updates the \emph{reached} flag of the motion module, and publishes to ROS topics for motion-related commands, such as waypoint or path following. It also provides the user the ability to use different path planning modules as long as they support the interface functions. \reffig{pathplanners} shows two agents executing the \emph{same application} using different path planners.\vspace{-1mm}

\subsection{Portability}

Apart from the dynamics, all aforementioned components of the \CyPhyHouse middleware are platform-agnostic.
Our implementation allows any agent or system simulating or deploying a \Koord program to use a configuration file (as shown in \reffig{config}) to specify the system configuration, and the runtime modules for each agent, including the dynamics-related modules, while using the same application code.

\section{CyPhyHouse multi-robot simulator}
\label{sec:simulator}
\begin{figure*}[h!]
\centering
\begin{subfigure}[b]{0.49\textwidth}
	\centering
	\includegraphics[width=\textwidth]{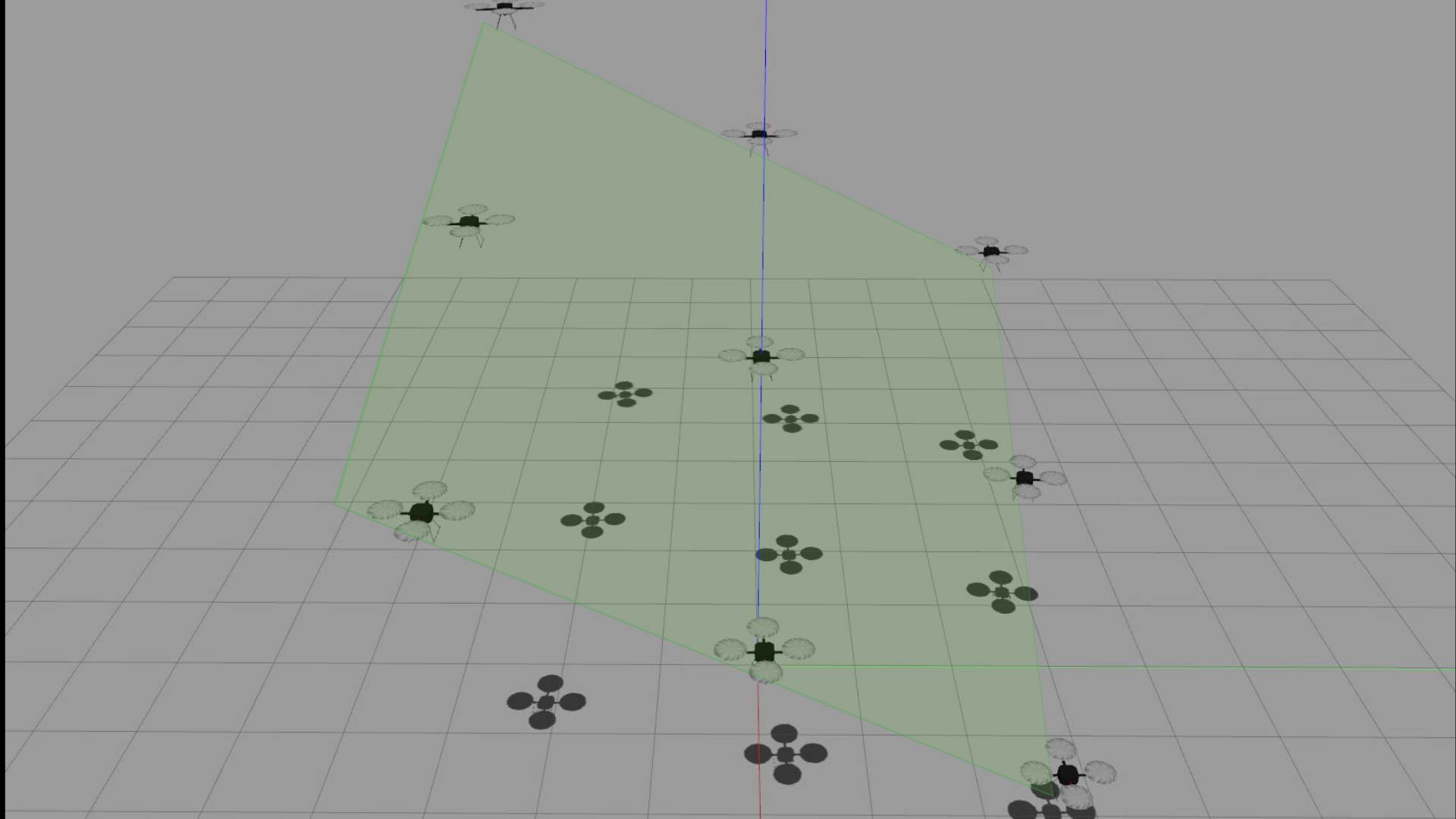}
	\caption{$\Shapeform$ with 9 drones.}
	\label{subfig:shape9}
\end{subfigure}
\begin{subfigure}[b]{0.49\textwidth}
	\centering
	\includegraphics[width=\textwidth]{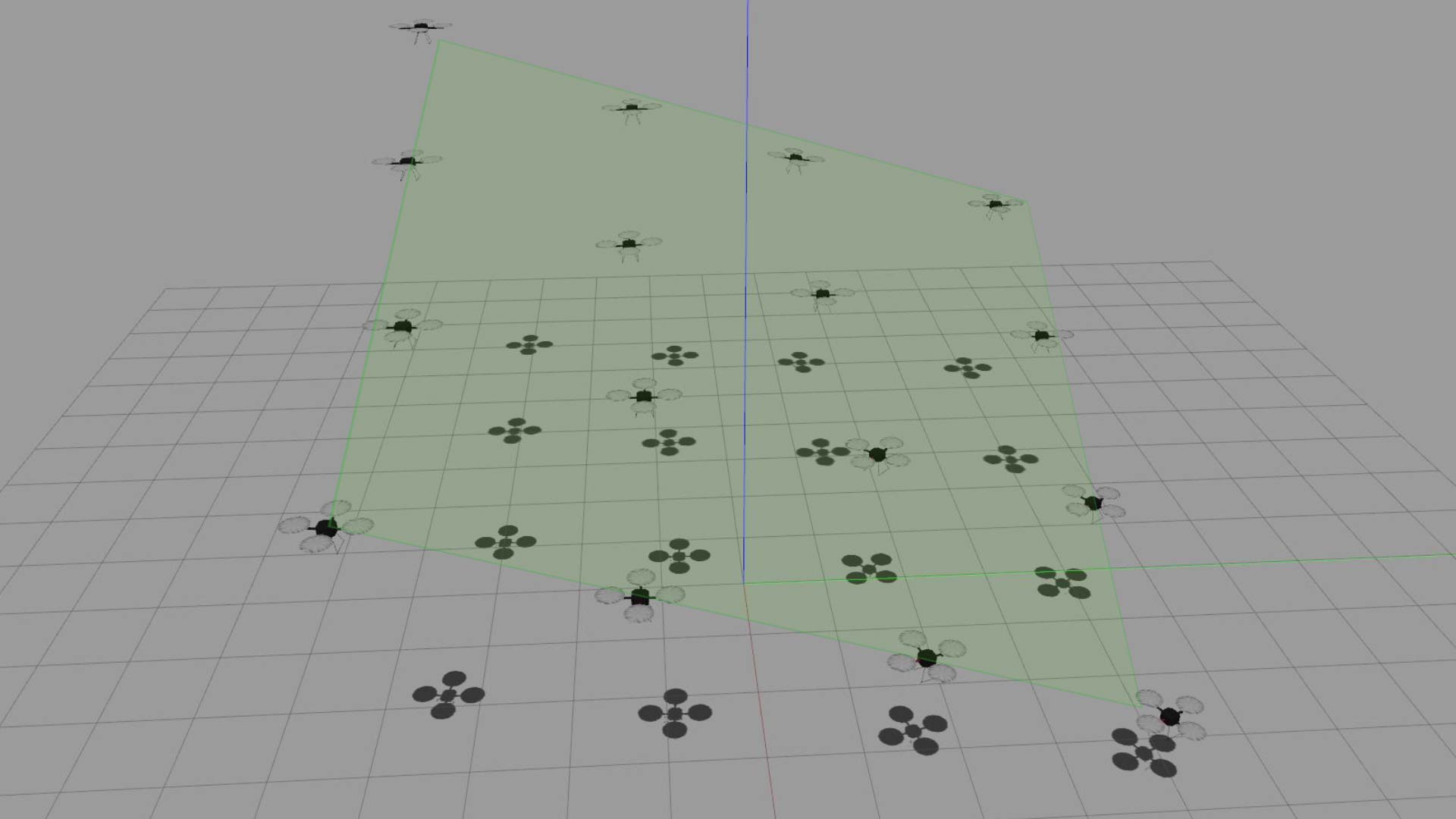}
	\caption{$\Shapeform$ with 16 drones.}
	\label{subfig:shap16}
\end{subfigure}
\begin{subfigure}[b]{0.49\textwidth}
	\centering
	\includegraphics[width=\textwidth]{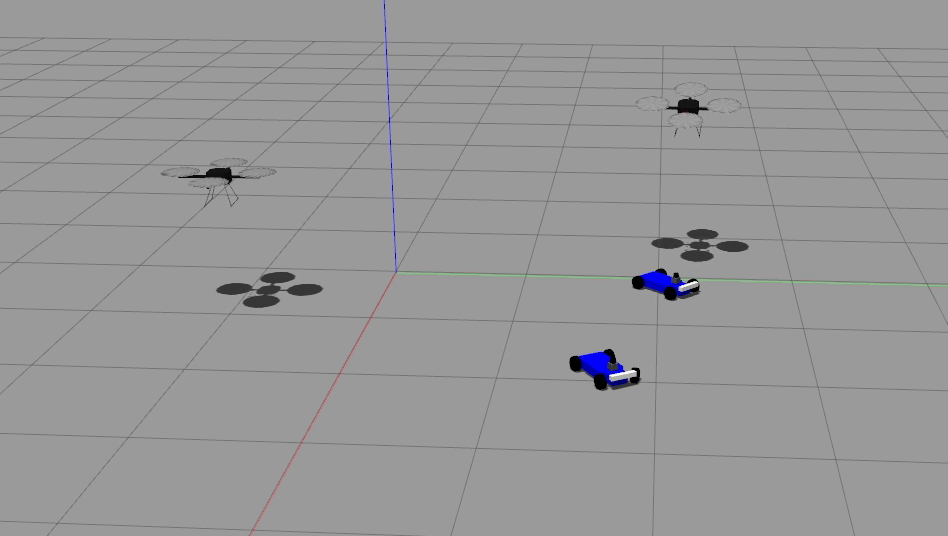}
	\caption{$\TaskApp$.}
	\label{subfig:rrt}
\end{subfigure}
  \caption{\CyPhyHouse\ simulator running different scenarios with the same \Koord application. \textbf{(a)} shows simulation of 9 drones running $\Shapeform$ application, \textbf{(b)} shows the $\Shapeform$ application on 16 drones. Different scenarios are specified by changing the configuration file. \textbf{(c)} shows a simulation of $\TaskApp$ on heteterogenous robots. }
  \label{fig:simexp}
\end{figure*}

We have built a high-fidelity simulator for testing distributed \Koord applications with large number of heterogeneous robots in different scenarios.
Our middleware design allows us to separate the simulation of \Koord applications and communications from the physical models for different platforms.
Consequently, the compiled \Koord applications together with the communication modules can run directly in the simulator---one instance for each participating robot, and only the physical dynamics and the robot sensors are replaced by their simulated counterparts. 
This flexibility enables users to test their \Koord applications under different scenarios and with various robot hardware platforms. Simpler physical models can be used for early debugging of algorithms; and the same code can be used later with more accurate physics and heterogeneous platforms.
The simulator can be used to test different scenarios, with different numbers of (possibly heterogeneous) robots,
with no modifications to the application code itself, rather simply modifying a configuration file as shown in \reffig{simexp}.
\sayan{To our knowledge, this is the only simulator for distributed robotics providing such fidelity and flexibility.}

\subsection{Simulator Design}
\paragraph*{Simulating \Koord and communication}
To faithfully simulate the communication, our simulator spawns a process for each robot which encompasses all middleware threads.
The communication handling threads in these processes can then send messages to each other through broadcasts within the local network.
To simulate robots on a single machine, we support specifying distinct network ports for robots in the configuration file.
Since the communication is through actual network interfaces, our work can be extended to simulate different network conditions with existing tools in the future.
\paragraph*{Physical Models and Simulated World}
Our simulated physical world is developed based on \Gazebo~\cite{gazebo}
and we provide a simulated positioning system to relay positions of simulated devices from \Gazebo to the \CyPhyHouse middleware.
We integrate two \Gazebo robot models from the \Gazebo and ROS community,
the car from the MIT RACECAR project~\cite{MIT_RACECAR} and the quadcopter from the hector quadrotor project~\cite{hector_quadrotor}.
Further, we implement a simplified version of position controller by modifying the provided default model.
Users can choose between simplified models for faster simulation or original models for accuracy.

In addition to simulation, we also develop Gazebo plugins for visualization.
Users may either use these to plot the movements or traces of the robots for real-time monitoring during experiments or visualize and analyze execution traces with Gazebo after experiments.
\subsection{Simulator performance analysis experiments and results.} 
Large scale simulations play an important role in testing robotic applications and also in training machine learning modules for perception and control.
Therefore, we perform a large set of experiments to measure the performance and scalability of the \CyPhyHouse\ simulator and experiment with various scenarios (such as different application \Koord programs,
increasing numbers of devices, or mixed device types). We then collect the usages of different resources and the amount of messages in each scenario.
Finally, we compare resource usages and communications to study how our simulator can scale across different scenarios.

In our experiments, we use three \Koord programs including the example \TaskApp in~\reffig{taskapp}, a line formation program \Lineform, and a program forming a square \Shapeform.
For \TaskApp, we simulate with both cars and quadcopters to showcase the coordination between heterogeneous devices.
\begin{figure}[h!]
\centering
    \includegraphics[width=0.7\textwidth]{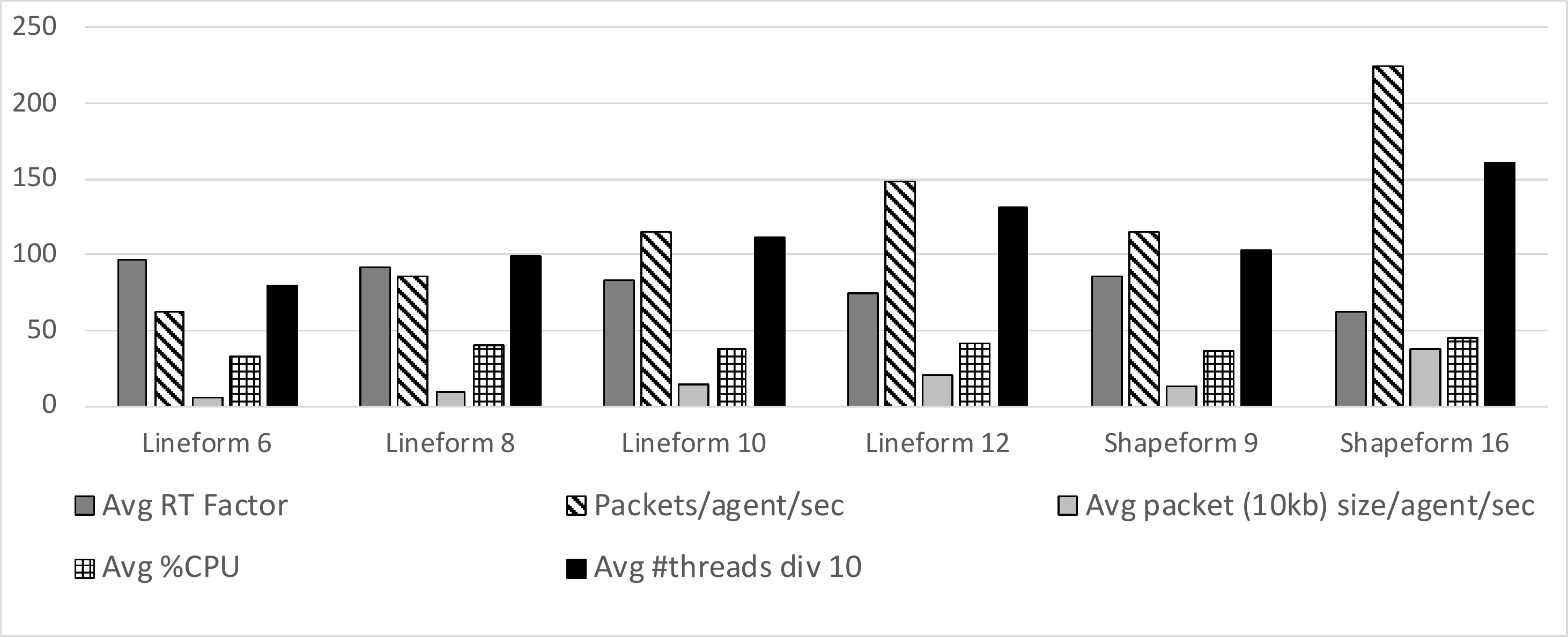}
    \vspace{0.1cm}
    \caption{Resource usages and communications for \Shapeform and \Lineform.}
    \label{fig:simstats}
\end{figure}
For \Lineform and \Shapeform, we use only quadcopters to evaluate the impact of increasing numbers of robots on these statistics.
For each experiment scenario with a timeout of 120 seconds, we collect the total message packets and packet length received by all robots
and sample the following resource usages periodically:
Real Time Factor (RT Factor, the ratio between simulated clock vs wall clock), CPU percentage, the memory percentage and the number of threads.
All experiments are run on a workstation with 32 Intel Xeon Silver 4110 2.10GHz CPU cores and 32 GB main memory.

In \reffig{simstats}, we only show the average of each collected metric.
For \Lineform and \Shapeform, RT factor drops while all resource usages scale linearly with the number of robots.
Average number and size of packets received per second for each robot also grows linearly;
hence, the message communication complexity for all robots is quadratic in the number of robots.
One can improve the communication complexity with a more advanced distributed shared memory design.

\section{Deployment Setup}
\label{sec:hardware}

\subsection{Vehicle platforms}

As previously mentioned, the \CyPhyHouse\ toolchain was developed with heterogeneous robotics platforms in mind.
In order to demonstrate such capabilities, we have built several cars and quadcopters.
Note that the cars have nonholonomic constraints, while the quadcopter has uncertain dynamics, so typically to deploy an application like \TaskApp, a roboticist would have to develop a separate application for each platform. With \CyPhyHouse we show that the same application code can be deployed on both platforms.

\subsubsection*{Quadcopters}
Two quadcopters were assembled from off-the-shelf hardware, with a $40 \text{cm} \times 40 \text{cm}$ footprint.
The main computer is a Raspberry Pi 3 B+ along with   a Navio2 deck for sensing and motor control. 
Stabilization and reference tracking are handled by Ardupilot~\cite{ardupilot}.
Between the \CyPhyHouse\ middleware and Ardupilot we include a hardware abstraction layer to convert setpoint messages from the high-level language into MAVLINK using the mavROS~\cite{mavros}.
Since the autopilot is designed for GPS, we  convert the current quadcopter position into the Geographic Coordinate System before sending it to the controller.

\subsubsection*{Car}

The cars are built based on the open-source MIT RACECAR project~\cite{MIT_RACECAR}.
The computing unit consists of an NVIDIA TX2 board. We wrote a custom ROS node that uses the current position and desired waypoints to compute the input speed and steering angle using a Model Predictive Controller (MPC). Low-level motor  control uses an electronic speed controller.
\begin{figure}[h!]
\centering 
  \includegraphics[width=0.6\textwidth]{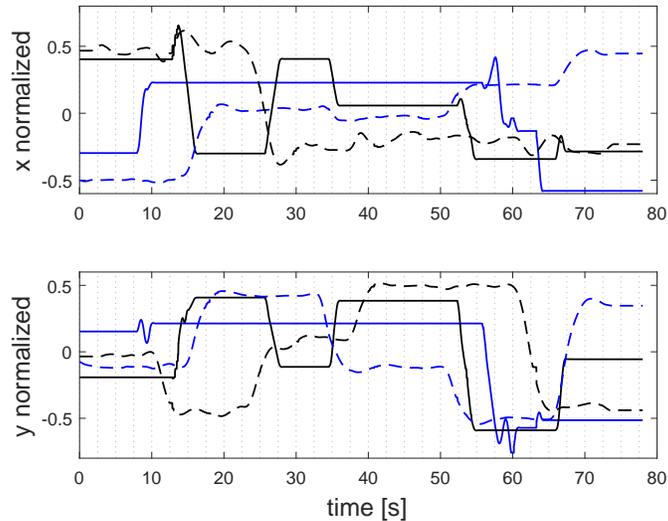}
  
  \caption{Normalized position traces for \TaskApp running on 2 quadcopters (dashed lines) and 2 cars (solid lines);  $x\ vs\ \mathit{time}$ (\emph{Top}) and $y\ vs\ \mathit{time}$ (\emph{Bottom}) traces. Note the concurrent safe movement of vehicles (e.g., around 13, 36, and 52 seconds). The $y$ positions of car 1 (solid black) and drone 2 (dashed blue) are close between 20-25s, but they are safe as they are separated in $x$.}
  \label{fig:taskdepl}
  \end{figure}
  
\subsection{Test arena and localization}
We performed our experiments in a $7 \text{m} \times 8 \text{m} \times 3 \text{m}$ arena equipped with 8 Vicon cameras.
The Vicon system allows us to track the position of multiple robots with sub-millimeter accuracy, however, we note that the position data can come from any source (for example GPS, ultrawide-band, LIDAR), as long as all robots share the same coordinate system.
%
%
While the motion capture system transmits all the data from a central computer, each vehicle only subscribes to its own position information.
This was done to simplify experiments, as the goal of this work is not to develop new positioning systems.
All coordination and de-conflicting across robots is performed based on position information shared explicitly through shared variables in the \Koord application. 

\subsection{Interface with middleware}

The same application can be deployed using different path planners and low-level controllers 
through interfaces defined by the \CyPhyHouse\ middleware (See~\refsect{middleware}). In our experiments, both vehicles use RRT-based path planners~\cite{lavalle1998rapidly}.  The planner uses a bicycle model for the cars, and a straight-line model for the quadcopters. 
%
The path generated is then forwarded to the robot via a ROS topic.
The relevant ROS topics for each vehicle were specified in the configuration. Each vehicle updates the \emph{reached} topic when it reaches the destination (\refsect{overview}).

\subsection{Experiments with \TaskApp on upto four vehicles}
The \TaskApp application of \refsect{overview} was run in over 100 experiments with different combinations of cars and quadcopters.  
\reffig{taskdepl} shows the $(x,y)$-trajectories of the vehicles in one specific trial run, in which two  quadcopters and two cars were deployed. 
Careful examination of such figures show us that all the performance requirements of \TaskApp are achieved: concurrent movement when different robots have clear paths to tasks, safe separation at all times, and robots getting blocked when there is no safe path found. In our experiments with up to $4$ vehicles, we found that with fewer robots, there are fewer blocked paths, so each robot spends less time idling. This non-blocking effect is superseded by the parallelism gains obtained from  having multiple robots. For example, with three robots (2 quadcopters and 1 car, or 1 quadcopter and 2 cars) \TaskApp had an average runtime of about 110 seconds for 20 tasks.
  The average runtime for the same with 4 robots across 70 runs was about 90 seconds. We experience zero failures, provided the wireless network conditions satisfies the assumptions stated in \refsect{middleware}.

\section{Conclusions}
\label{sec:conclusion}
We presented the open source \Koord language and the \CyPhyHouse\ toolchain for distributed robotic applications. 
We demonstrated the usefulness of the language in succinctly writing an application involving distributed task allocation and path planning. We presented and profiled the scalable \CyPhyHouse simulator that can  execute and test instances of \Koord applications with dozens of vehicles. We also showed how the {\em same\/} code can be simulated, and directly deployed on cars and drones with supporting platform specific controllers. While still in the development stages, \CyPhyHouse has been used by more than 25 individuals for programming, simulating, and testing other applications like formation flight, and surveillance. Our experiences suggest the toolchain can indeed lower the barrier for entry into the distributed robotics.

\subsection*{Acknowledgements}
We would like to thank Sasa Misailovic for contributing to the design of the Koord language. 
John Wang and Zitong Chen contributed to \Koord and \CyPhyHouse software systems.
We are grateful to Nitin Vaidya for the early design of the shared memory system. 
This research is supported by a research grant from the National Science Foundation (CNS award number 1629949).
%
\small
\bibliographystyle{unsrt}
\bibliography{ms}

\end{document}